%% file: DFWSGAR.tex
\documentclass[10pt,twocolumn,letterpaper]{article}

\usepackage[pagenumbers]{cvpr} %

\usepackage{graphicx}
\usepackage{amsmath}
\usepackage{amssymb}
\usepackage{booktabs}
\usepackage{xcolor, colortbl}
\usepackage{verbatim}
\usepackage{tabularx}
\usepackage{pifont}     %
\usepackage{array}
\usepackage{cuted}
\usepackage[inline]{enumitem}
\usepackage[accsupp]{axessibility}  %

\usepackage[pagebackref,breaklinks,colorlinks]{hyperref}
 
\usepackage[capitalize]{cleveref}
\crefname{section}{Sec.}{Secs.}
\Crefname{section}{Section}{Sections}
\Crefname{table}{Table}{Tables}
\crefname{table}{Tab.}{Tabs.}

\definecolor{lblue}{rgb}{0.1, 0.2, 0.7}
\definecolor{orange}{rgb}{1.0, 0.5, 0.0}
\definecolor{purple}{rgb}{0.5, 0.2, 0.9}
\definecolor{Gray}{gray}{0.9}

\newcommand{\expnum}[2]{{#1}\mathrm{e}{-#2}}
\newcommand{\splitcell}[1]{\begin{tabular}{@{}c@{}}#1\end{tabular}}
\newcommand{\bsplitcell}[1]{$\left[\splitcell{#1}\right]$}
\newcommand{\beginsupplement}{%
        \setcounter{section}{0}
        \renewcommand{\thesection}{\Alph{section}}
        \setcounter{table}{0}
        \renewcommand{\thetable}{A\arabic{table}}%
        \setcounter{figure}{0}
        \renewcommand{\thefigure}{A\arabic{figure}}%
     }

\def\ie{\emph{i.e.}}
\def\eg{\emph{e.g.}}
\def\etal{\emph{et al.}}

\def\cvprPaperID{3946} %
\def\confName{CVPR}
\def\confYear{2022}

\begin{document}

\title{Detector-Free Weakly Supervised Group Activity Recognition}

\author{
Dongkeun Kim$^1$
\qquad
Jinsung Lee$^2$
\qquad
Minsu Cho$^{1,2}$
\qquad
Suha Kwak$^{1,2}$\\
Department of CSE, POSTECH$^1$ \qquad Graduate School of AI, POSTECH$^2$\\
\small{\url{https://cvlab.postech.ac.kr/research/DFWSGAR/}}
}

\maketitle

\input{_0_abstract_arxiv}

\input{_1_introduction_arxiv}

\input{_2_related_work_arxiv}

\input{_3_method_arxiv}

\input{_4_experiments_arxiv}

\input{_5_conclusion_arxiv}

\vspace{1mm}
{\small
\noindent \textbf{Acknowledgement.} 
This work was supported by 
the NRF grant and       %
the IITP grant          %
funded by Ministry of Science and ICT, Korea
(NRF-2021R1A2C3012728,  %
 NRF-2018R1A5A1060031,  %
 IITP-2020-0-00842,     %
 IITP-2021-0-00537,     %
 No.2019-0-01906 Artificial Intelligence Graduate School Program--POSTECH).     %
}

{\small
\bibliographystyle{ieee_fullname}
\bibliography{cvlab_gar}
}

\clearpage
\beginsupplement

\noindent
\textbf{\Large Appendix}

\input{_supp_1_experiments_arxiv}

\input{_supp_2_qualitative_arxiv}

\end{document}

%% file: _0_abstract_arxiv.tex
\begin{abstract}
    Group activity recognition is the task of understanding the activity conducted by a group of people as a whole in a multi-person video. 
    Existing models for this task are often impractical in that they demand ground-truth bounding box labels of actors even in testing or rely on off-the-shelf object detectors.
    Motivated by this, we propose a novel model for group activity recognition that depends neither on bounding box labels nor on object detector. 
    Our model based on Transformer localizes and encodes partial contexts of a group activity by leveraging the attention mechanism, and represents a video clip as a set of partial context embeddings.
    The embedding vectors are then aggregated to form a single group representation that reflects the entire context of an activity while capturing temporal evolution of each partial context.    
    Our method achieves outstanding performance on two benchmarks, Volleyball and NBA datasets, surpassing not only the state of the art trained with the same level of supervision, but also some of existing models relying on stronger supervision. 
\end{abstract}

%% file: _1_introduction_arxiv.tex
\section{Introduction}

Group activity recognition (GAR) is the task of classifying the activity that a group of people are doing as a whole in a given video clip. 
It has attracted increasing attention due to a variety of its applications including sports video analysis, video surveillance, and social scene understanding. 
Unlike the conventional action recognition that focuses on understanding individual actions~\cite{simonyan2014two, tran2015learning, wang2016temporal, carreira2017quo, wang2018non, kwon2020motionsqueeze, fan2018end, repflow2019, girdhar2019video, lin2019tsm}, GAR demands comprehensive and precise understanding of interactions between multiple actors, which introduces inherent challenges such as localization of actors and modeling their spatio-temporal relations.

Due to the difficulty of the task, most of existing methods for GAR~\cite{ibrahim2016hierarchical, wu2019learning, hu2020progressive, gavrilyuk2020actor, pramono2020empowering, ehsanpour2020joint, yan2020higcin, yuan2021learning, li2021groupformer} require ground-truth bounding boxes of individual actors for both training and testing, and their action class labels for training. 
In particular, the bounding box labels are used to extract features of individual actors  (\eg, \emph{RoIPool}~\cite{faster_rcnn} and \emph{RoIAlign}~\cite{mask_rcnn}) and discover their spatio-temporal relations precisely; such actor features are aggregated while considering the relations between actors to form a group-level video representation, which is in turn fed to a group activity classifier. 
Though these methods have demonstrated impressive performance on the challenging task, their dependence on the heavy annotations, especially bounding boxes at inference, is impractical and in consequence restricts their applicability significantly.

One way to resolve this issue is to jointly learn group activity recognition and person detection using bounding box labels~\cite{bagautdinov2017social, zhang2019fast} to estimate bounding boxes of actors at inference. 
This approach however still requires ground-truth bounding boxes of individual actors for training videos.
To further reduce the annotation cost, Yan~\etal~\cite{yan2020social} introduced weakly supervised GAR (WSGAR) that does not demand actor-level labels at both training and inference. 
They address the lack of bounding box labels by generating actor box proposals through a detector pretrained on an external dataset and learning to prune irrelevant proposals.

\begin{figure}[t]
\centering
\includegraphics[width=1.0\linewidth]{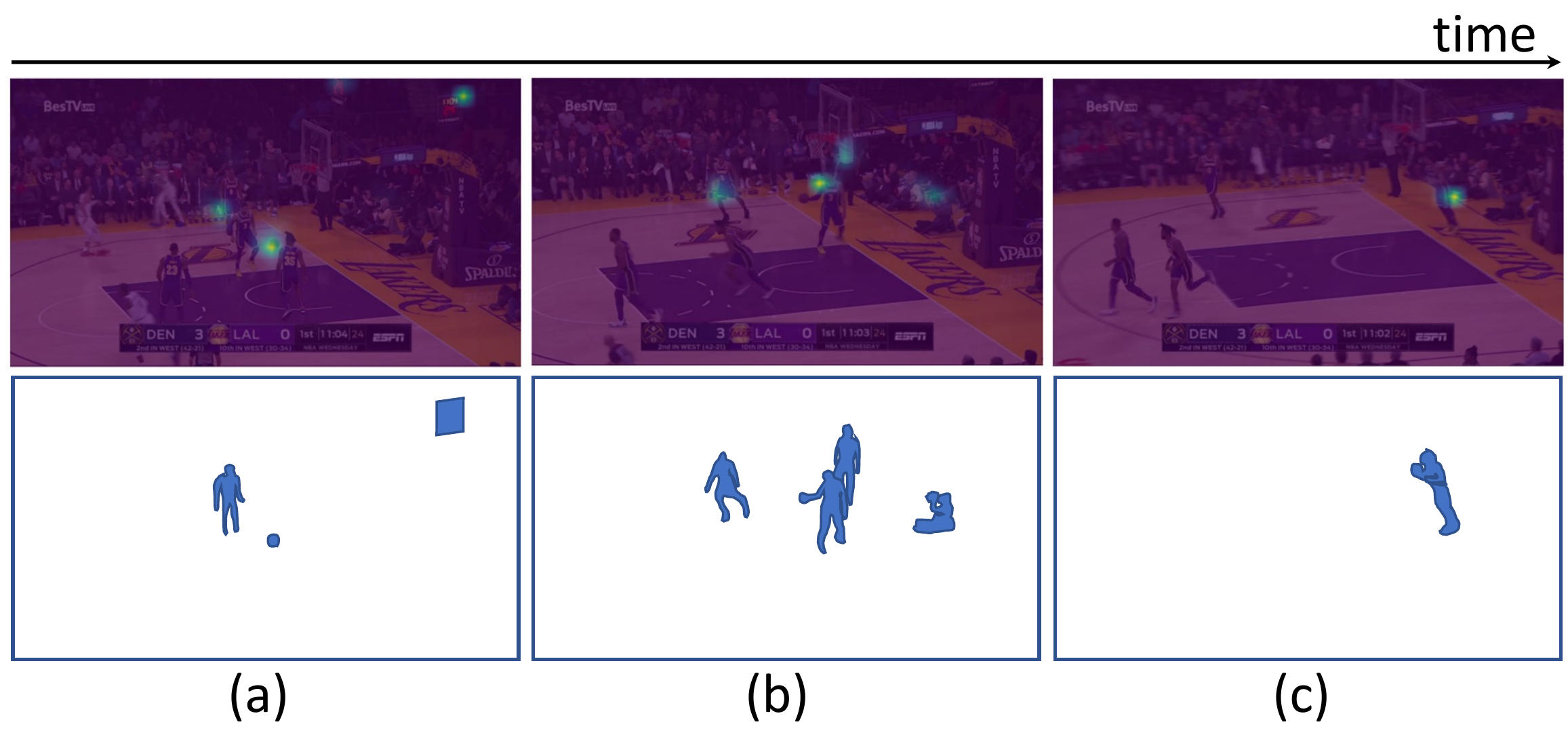}
\vspace{-1.5em}
\caption{
Visualization of partial contexts captured by a token across time. The token in this example focuses on how players behave after they conceded a goal.
(a) Right after the event \textit{3p-succ}, the timer has been reset to 24 secs and a defender stares at the ball. (b) Players prepare for the next attack, while a referee and a cameraman point at who takes the ball. (c) A player initiates the next attack. With such tokens each representing different pieces of the whole group activity, our model acquires the encapsulated semantics of the target activity.
}
\label{fig:fig1}
\vspace{-1.0em}
\end{figure}

The detector-based WSGAR however has several drawbacks as follows.
First of all, a detector often suffers from occlusion and background clutter, and thus frequently causes missing and false detections that degrade GAR accuracy. 
Second, the detector-based approach loses contextual information that are useful for GAR since it concentrates only on people;
in sports video analysis, for example, entities other than people, such as a ball and a scoreboard, may provide crucial information for the task. 
Third, object detection is costly to itself and imposes additional overheads in both computation and memory.

In this paper, we propose a detector-free model for WSGAR that depends neither on ground-truth bounding boxes nor on object detector.
It bypasses explicit object detection by drawing attention on entities involved in a group activity through a Transformer encoder~\cite{vaswani2017attention} placed on top of a convolutional neural network (CNN) backbone.
Specifically, 
we define learnable tokens as input to the encoder so that each of them learns to localize partial contexts of a group activity through the attention mechanism of the encoder; the tokens capture not only key actors but also other useful clues as shown in Fig.~\ref{fig:fig1}.
Since a set of learnable tokens are shared for all frames, a predefined number of token embeddings are computed by the encoder for every frame.
A video clip is then represented as a bag of token embeddings, which are aggregated into a group representation in two steps:
Those computed from the same token at different frames are first aggregated to capture the temporal evolution of each token, then the results are fused to form a single feature vector for group activity classification. 

In addition, for further performance improvement, the backbone of our model is designed to compute motion-augmented features.
Unlike previous work on GAR~\cite{azar2019convolutional, gavrilyuk2020actor, pramono2020empowering, li2021groupformer}, it does not rely on off-the-shelf optical flow that is prohibitively expensive and thus has been a computational bottleneck.
Instead, inspired by recent video representation architectures~\cite{kwon2020motionsqueeze,kwon2021learning,fan2018end,repflow2019, wang2020video}, 
it learns to capture motion information in feature levels by embedding local correlation between the feature maps of two adjacent frames.

We evaluate the proposed framework on two datasets, Volleyball~\cite{ibrahim2016hierarchical} and NBA~\cite{yan2020social}. 
Our framework achieves the state-of-the-art performance on the two benchmarks in the weakly supervised learning setting, 
and is as competitive as existing methods relying on stronger supervision such as ground-truth bounding boxes and individual action class labels.
The contribution of this paper is three-fold: 
\vspace{-2mm}
\begin{itemize}[leftmargin=5mm] 
\itemsep=-0.5mm
    \item 
    We present the first detector-free method dedicated to WSGAR, which demands neither ground-truth bounding box labels nor object detector.
    \item We propose a novel Transformer-based model that captures key actors and objects involved in a group activity 
    through the attention mechanism.
    Moreover, our model is carefully designed to capture their temporal dynamics to produce a rich group-level video feature.
    \item On the two benchmarks, the proposed method largely outperforms existing WSGAR models. Also, it even beats early GAR models that depend on stronger supervision than ours.
\end{itemize}

%% file: _2_related_work_arxiv.tex
\section{Related work}

\begin{figure*}[!t]
\begin{center}
\includegraphics[width=1.0\linewidth]{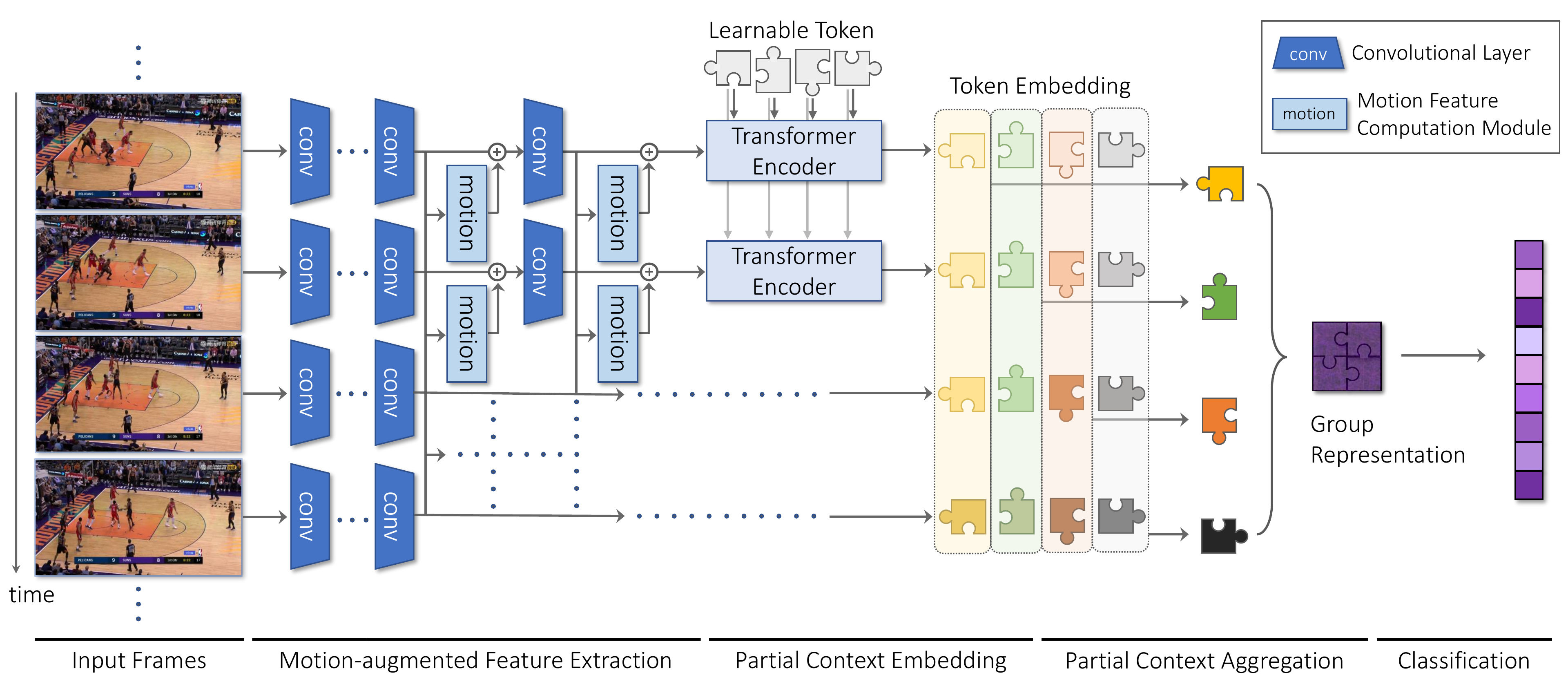}
\end{center}
\vspace{-5mm}
\caption{
Overall architecture of our model. A CNN incorporating motion feature computation modules extracts a motion-augmented feature map per frame. 
At each frame, a set of learnable tokens (unpainted pieces of Jigsaw puzzles) are embedded to localize clues useful for group activity recognition through the attention mechanism of the Transformer encoder. 
The token embeddings (painted pieces of Jigsaw puzzles) are then fused to form a group representation in two steps: First aggregate embeddings of the same token (pieces with the same shape) across time, and then aggregate the results of different tokens (pieces with different shapes and colors).
Finally, the group representation is fed into the classifier which predicts group activity class scores. 
}
\label{fig:fig2}
\vspace{-3mm}
\end{figure*}

\subsection{Group activity recognition}

On account of its various applications in the real world, group activity recognition (GAR) has been studied extensively. 
Earlier attempts employ hand-crafted features with probabilistic graphical models~\cite{choi2009they, ryoo2011stochastic, choi2012unified, lan2012social, amer2014hirf} or AND-OR graphs~\cite{amer2012cost, amer2013monte, shu2015joint} based on inherent relationship between individual actions and group activities.
RNN-based methods~\cite{bagautdinov2017social, deng2016structure, ibrahim2018hierarchical, ibrahim2016hierarchical, li2017sbgar, qi2018stagnet, shu2017cern, wang2017recurrent, yan2018participation} have shown the effectiveness of hierarchical temporal modeling.
LSTM architecture is often structured in hierarchical ways~\cite{ibrahim2016hierarchical, shu2017cern, wang2017recurrent} to model individual action dynamics and aggregate individual features to infer group activity. 
Moreover, graphically constructed RNN models~\cite{deng2016structure, qi2018stagnet} are proposed to utilize relationship among the individual features.

Recent approaches show more tendency towards adopting relational modeling~\cite{azar2019convolutional, wu2019learning, ehsanpour2020joint, hu2020progressive, yan2020higcin, yuan2021spatio}. 
Graph-based approaches have been widely used to model spatio-temporal relationship between actors~\cite{wu2019learning, ehsanpour2020joint, hu2020progressive, yan2020higcin, yuan2021spatio}; they first extract features from bounding boxes, and then place features as nodes and defining their relations as edges. 
These approaches subsequently employ their own way of evolving relation graphs, such as graph convolutional networks (GCN)~\cite{wu2019learning} or graph attention networks (GAT)~\cite{ehsanpour2020joint}. 
More complicated ways to develop relation graphs, for instance, constructing cross inference module to embed spatio-temporal features~\cite{yan2020higcin} or utilizing dynamic relation and dynamic walk offsets to build person-specific interaction graph~\cite{yuan2021spatio}, have also been introduced.  
On the other hand, Azar~\etal~\cite{azar2019convolutional} introduce the notion of activity map to encode spatial relations between individuals.

Transformer-based methods~\cite{gavrilyuk2020actor, pramono2020empowering, pramono2021relational, li2021groupformer, yuan2021learning} model the relationship between features of group activities, and show significant improvements in GAR. 
They place a Transformer 
on top of the actor features to embed spatio-temporal relational contexts with conditional random fields (CRF)~\cite{pramono2020empowering} or joint spatio-temporal contexts regarding intra- and inter-group relations~\cite{li2021groupformer}. 
The most related study by Yuan \etal~\cite{yuan2021learning}  encodes person-specific scene context per individual feature. 
However, it still relies on person detector and only captures person-specific context. 
On the other hand, our method further considers multiple people and does not depend on any off-the-shelf detector, and thus enables the model to be trained effectively with less supervision. 
Moreover, ours utilizes learnable tokens to form partial context shared by different group activities.

\noindent
\textbf{Weakly supervised group activity recognition.}
GAR has many obstacles to overcome in order to be applicable in real life.
In particular, heavy annotations such as bounding boxes and individual actions are rarely provided.
Thus, several methods have addressed GAR with weaker supervision such as utilizing bounding boxes only to train their built-in detector~\cite{bagautdinov2017social, zhang2019fast} or activity map~\cite{azar2019convolutional}. 
Accordingly, Yan~\etal~\cite{yan2020social} propose WSGAR, the task where bounding box annotations are not used in both training and inference. They address the absence of the bounding boxes by placing an off-the-shelf object detector inside the model.
To prune off noisy outputs of the object detector, a relation graph is constructed regarding the relatedness of detected bounding boxes. 
Zhang~\etal~\cite{zhang2021multi} propose a method using activity-specific features for multi-label activity recognition, which also shows an improvement in WSGAR. 
However, it is not designed for GAR and the performance gap from baseline is subtle in the WSGAR setting.
Unlike the previous work~\cite{yan2020social,zhang2021multi}, we propose a detector-free method dedicated to WSGAR, which is not only free from actor-level annotation but also from object detector.

\subsection{Transformer}
Transformer~\cite{vaswani2017attention} is originally devised to solve sequence-to-sequence task such as machine translation. 
It introduces self-attention mechanism with the aim of capturing global dependencies of the input elements. 
Recently, Transformer is widely adopted in many vision tasks either with a CNN feature extractor~\cite{carion2020end, meinhardt2021trackformer, zou2021end, kim2021hotr, girdhar2019video} or as a pure transformer architecture~\cite{dosovitskiy2021an}. 
Detection Transformer (DETR)~\cite{carion2020end}, a Transformer model for object detection, is presented and utilized in detection-based tasks such as human-object interaction detection~\cite{kim2021hotr, zou2021end} and object tracking~\cite{meinhardt2021trackformer}. 
Moreover, several attempts apply self-attention mechanism to process videos~\cite{wang2018non, bertasius2021space, girdhar2019video, ryoo2021tokenlearner}.
Girdhar~\etal~\cite{girdhar2019video} utilize spatio-temporal context around the person to localize and recognize human actions simultaneously. 
Ryoo~\etal~\cite{ryoo2021tokenlearner} propose TokenLearner, which efficiently learns to convey meaningful features of an input.

%% file: _3_method_arxiv.tex
\section{Proposed method}

\begin{figure*}
     \centering
     \begin{subfigure}[b]{0.34\textwidth}
         \centering
         \includegraphics[width=\textwidth]{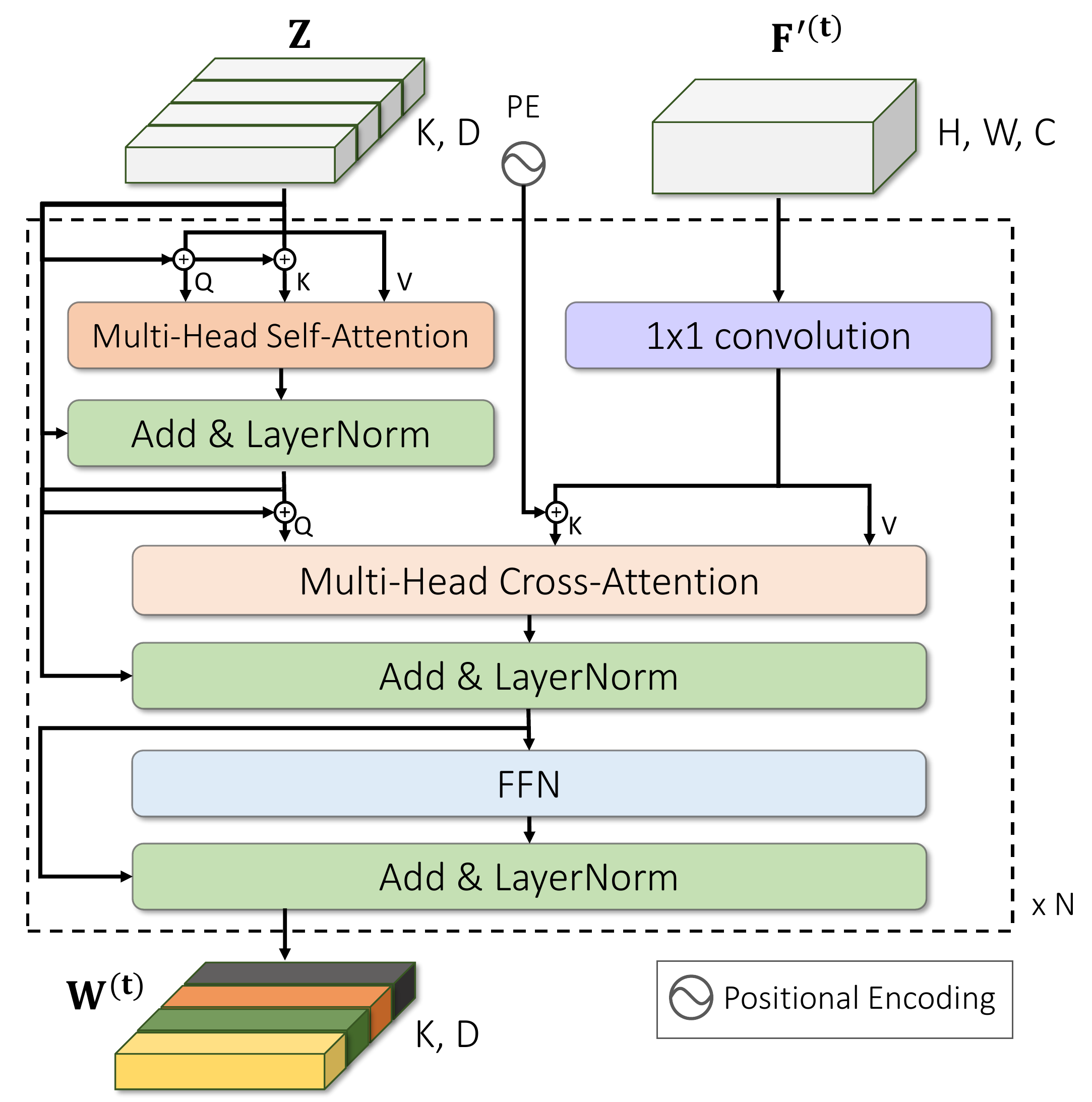}
         \caption{Partial Context Embedding}
         \label{fig:PCE}
     \end{subfigure} \hspace{1mm}
     \begin{subfigure}[b]{0.63\textwidth}
         \centering
         \includegraphics[width=\textwidth]{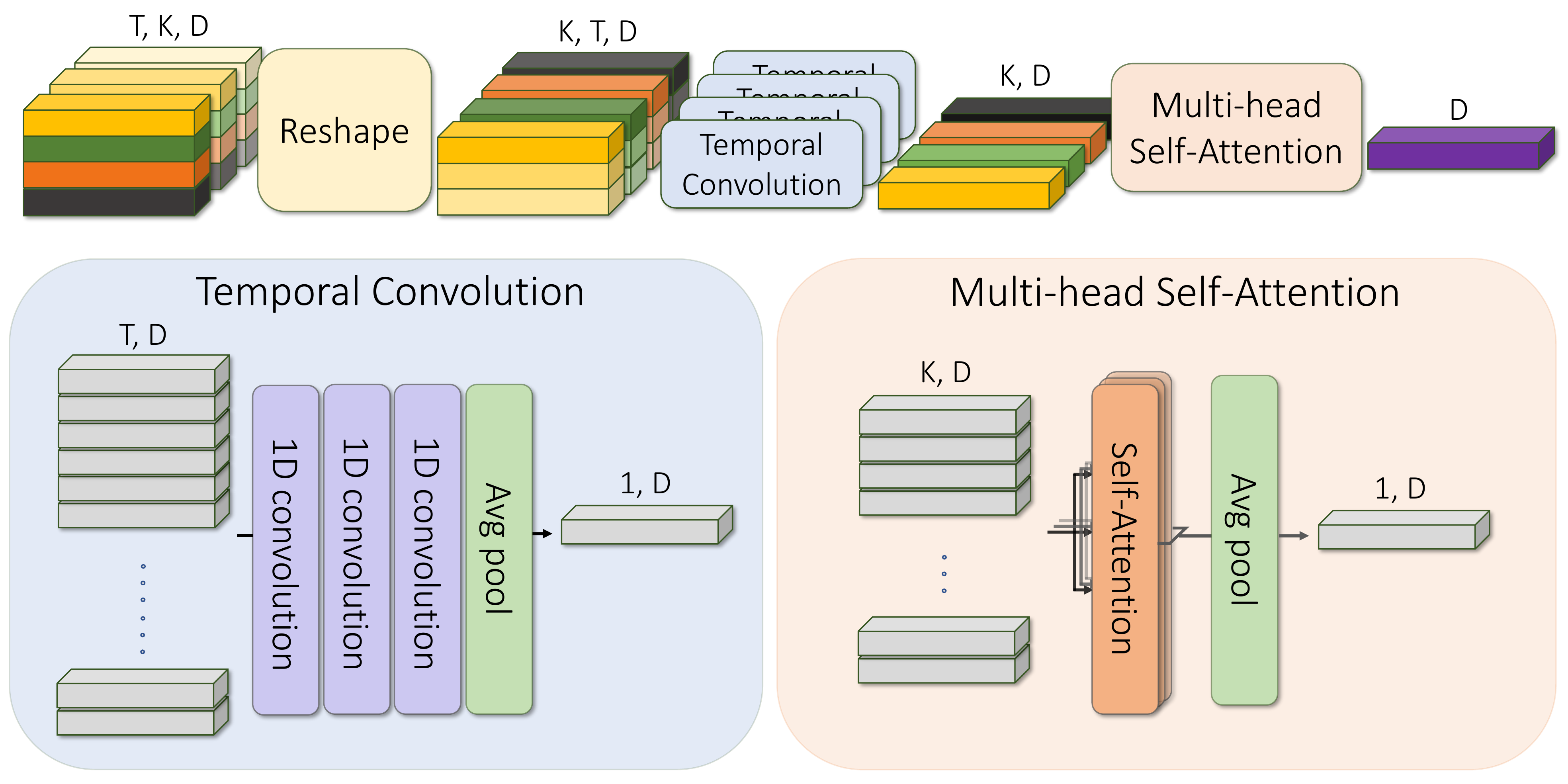}
         \caption{Partial Context Aggregation}
         \label{fig:PCA}
     \end{subfigure}
     \vspace{-2mm}
     \caption{Detailed architectures of the partial context embedding and partial context aggregation modules.}
     \vspace{-2mm}
\end{figure*}

Our goal is to recognize a group activity in a multi-person video without using ground-truth bounding boxes or object detector. 
We achieve this goal by leveraging the attention mechanism to localize and encode partial contexts of a group activity, and then aggregating them into a group-level video representation while capturing their temporal dynamics.
Our model is divided into three parts: motion-augmented feature extraction, partial context embedding, and partial context aggregation.
Its overall architecture is illustrated in Fig.~\ref{fig:fig2}, and the remainder of this section elaborates on each of the three parts.

\subsection{Motion-augmented feature extraction}
Given a video clip of $T$ frames $\textbf{X}_{video} \in \mathbb{R}^{T \times H^{0} \times W^{0} \times 3}$ as input, an ImageNet pretrained ResNet~\cite{resnet} backbone extracts features
$\textbf{F}_{video} \in \mathbb{R}^{T \times H \times W \times C}$ in a frame-wise manner.
To incorporate motion information into the features without using computationally heavy 3D CNNs~\cite{tran2015learning, carreira2017quo} or optical flows~\cite{zach2007duality, sun2018pwc}, our model computes local correlation between two adjacent intermediate feature maps, $\textbf{F}^{(t)}$ and $\textbf{F}^{(t+1)}$, 
and encodes the correlations into frame-wise motion features, similar to recent motion feature learning methods~\cite{wang2020video, kwon2020motionsqueeze}. 

\noindent
\textbf{Motion feature computation.}
Given two feature maps of adjacent frames, 
we first reduce their channel dimension to 
$C^{\prime}$ by $1\times1$ convolution.
Then, the local correlation function $s$ is defined as follows:

$$s: (\textbf{F}^{(t)}, \textbf{F}^{(t+1)}) \longmapsto \textbf{S}^{(t)} \in \mathbb{R}^{H \times W \times P \times P}$$
where
\begin{equation}
    (\textbf{S}^{(t)})_{(\textbf{x}, \textbf{p})} = \langle (\textbf{F}^{(t)})_{(\textbf{x})} ,(\textbf{F}^{(t+1)})_{(\textbf{x}+\textbf{p})}\rangle, 
\label{eqn:eq1}
\end{equation}
$\textbf{x} \in [0,H-1] \times [0,W-1]$, and $\textbf{p} \in [-l, l]^2$. 
An element $(\textbf{S}^{(t)})_{(\textbf{x},\textbf{p})}$ of the local correlation tensor $\textbf{S}^{(t)}$ is calculated by the dot product similarity of displaced vectors between adjacent frames $\textbf{F}^{(t)}$ and $\textbf{F}^{(t+1)}$.
By restricting the maximum displacement to $l$, the correlation scores of spatial position \textbf{x} are computed only in its local neighborhood of size $P=2l+1$. 
Note that each feature map is zero-padded with the size of $l$. 
Hence, $\textbf{S}^{(t)}$ reveals the motion at each location of $\textbf{F}^{(t)}$ in the form of $P\times P$ local correlation map.
In the case of the $T$-th frame, $\textbf{S}^{(T)}$ is computed by the self-correlation $s(\textbf{F}^{(T)}, \textbf{F}^{(T)})$.
To integrate the local correlation tensor into the backbone, 
$1\times1$ convolution transforms the local correlation tensor 
$\textbf{S}^{(t)} \in \mathbb{R}^{H \times W \times P^2}$ 
into motion features $\textbf{M}^{(t)} \in \mathbb{R}^{H \times W \times C}$.
Then the motion features are inserted into the backbone using a residual connection, \ie,
$\textbf{F}^{\prime (t)} = \textbf{F}^{(t)} + \textbf{M}^{(t)}$; this operation endows the output features with the sense of motion.
In our model, two motion feature computation modules are inserted after the last two residual blocks of ResNet. 
Following~\cite{kwon2020motionsqueeze}, we adopt the local correlation computation implemented in FlowNet~\cite{dosovitskiy2015flownet}.

\subsection{Partial context embedding (PCE)}
Given the motion-augmented feature $\textbf{F}^{\prime} \in \mathbb{R}^{T \times H \times W \times C}$ for $T$ frames, a set of $K$ learnable tokens $\textbf{Z}=\{\textbf{z}_i\}_{i=1}^{K}$, 
where $\textbf{z}_i \in \mathbb{R}^D$ for all $i$,
is trained to encode partial contexts of a group activity through 
the Transformer encoder in a frame-wise manner.
For each frame, $K$ tokens are transformed into token embeddings $\textbf{W}^{(t)}$ of the same size using the mechanism of the encoder, that is,
the set of learnable tokens $\textbf{Z}$ is shared by every frame to capture temporal dynamics of each token embeddings across time.
To this end, we adopt a Transformer architecture~\cite{vaswani2017attention} composed of multi-head cross-attention layer, multi-head self-attention layer, and feed forward network (FFN).
The tokens are embedded while considering the relations with other tokens through the self-attention, and capture the partial contexts from the motion-augmented features through the cross-attention.
We call this process \emph{partial context embedding}, and its implementation follows the decoder of DETR~\cite{carion2020end}.
Note that the weights of the Transformer encoder are shared across $T$ frames.
For clarity, we describe the detailed process of partial context embedding for a frame $t$
given motion-augmented feature $\textbf{F}^{\prime (t)} \in \mathbb{R}^{H \times W \times C}$ and a set of learnable tokens $\textbf{Z}$ (Fig.~\ref{fig:PCE}). 

First, a point-wise convolution operation is applied to the feature map to reduce the channel dimension $C$ to $D$.
Then the spatial dimension of the feature map is flattened to convert its overall shape to $HW \times D$. 
For the multi-head cross-attention layer, the \textit{Query} is the sum of two elements: 
(1) the learnable tokens $\textbf{Z}$, and
(2) the output of the multi-head self-attention applied to the learnable token $\textbf{Z}$ and 
the output token embeddings of the previous encoder layer. 
The \textit{Key} and \textit{Value} come from the flattened feature map, and the spatial positional encoding is added to the \textit{Key}. 
To be specific, the spatial coordinates of the input feature map $\textbf{F}^{\prime (t)}$ are transformed into the spatial positional encoding using the sinusoidal function~\cite{vaswani2017attention};
each half of D channels are encoded with width and height coordinates, respectively.
Through the attention mechanism, each token learns to localize and encode partial contexts including key actors and entities from the given feature maps.  
For $T$ frames, the output token embeddings $\textbf{W}^{(t)} = \{\textbf{w}_i^{(t)}\}_{i=1}^{K}$ are stacked to form $\textbf{W} = [\textbf{W}^{(1)}, \textbf{W}^{(2)}, \ldots, \textbf{W}^{(T)}] \in \mathbb{R}^{T \times K \times D}$, a bag of $T \times K$ token embeddings of dimension $D$.
Note that $\textbf{w}_i^{(t)}$ denotes the $i$-th token embedding for a frame $t$.

\subsection{Partial context aggregation (PCA)}

The partial context aggregation module takes the output token embeddings $\textbf{W} \in \mathbb{R}^{T \times K \times D}$ and aggregates them to the final group representation $\textbf{g} \in \mathbb{R}^{D}$. 
This process is divided into two steps.
The first step aggregates embeddings of the same token across time and the second step assembles the results from the first step to build a single video representation. 
The overall process is illustrated in Fig.~\ref{fig:PCA}, and discussed in detail below.

In the first step, the output token embeddings $\textbf{W}_{i} = \{\textbf{w}_i^{(t)}\}_{t=1}^{T} \in \mathbb{R}^{T \times D}$ from the $i$-th token at different frames are fused to form an $i$-th aggregated token feature $\tilde{\textbf{w}_i} \in \mathbb{R}^{D}$ for all $i$.
After reshaping the output token embedding \textbf{W} to $K \times T \times D$ tensor, a stack of 1D convolutional layers, each followed by ReLU~\cite{relu}, is applied along $T$ dimension. 
The output is then fed to $AvgPool$ operation. 
This temporal convolution block $f(\cdot)$ gradually aggregates the token embedding along the time dimension, 
\ie, $f: \mathbb{R}^{K \times T \times D} \rightarrow \mathbb{R}^{K \times D}$. 
It achieves robustness to a temporal shift of activity in the video via parameter sharing in time while effectively capturing the temporal dynamics of token embeddings via stacked layers. 
In the second step, $K$ aggregated token features $\tilde{\textbf{W}} = \{\tilde{\textbf{w}}_i\}_{i=1}^{K} \in \mathbb{R}^{K \times D}$ are fused to form the group representation $\textbf{g}$. 
Specifically, layer normalization~\cite{layernorm} is first applied to $\tilde{\textbf{W}}$, %
and then a single layer multi-head self-attention is adopted to capture the dynamic relations between the $K$ aggregated token embeddings.
Finally, the group representation \textbf{g} is obtained by applying $AvgPool$ operation over $K$ dimension.

\subsection{Training objective}
After obtaining the group representation, a classifier is applied to predict group activity class scores. 
Our model is trained with the standard cross-entropy loss in an end-to-end manner.

%% file: _4_experiments_arxiv.tex
\section{Experiments}

We evaluate the proposed detector-free model for WSGAR on two datasets, Volleyball~\cite{ibrahim2016hierarchical} and NBA~\cite{yan2020social},
where our model is compared with the state-of-the-art WSGAR and GAR methods. 
We also validate the effectiveness of the model by extensive ablation studies and qualitative analysis. 

\subsection{Datasets}

\noindent
\textbf{Volleyball dataset.}
This dataset consists of 55 videos, which are further divided into 4830 clips. 3494 clips among them are used for training and 1337 clips are kept for testing. 
The center frame of each clip is labeled with 
\begin{enumerate*}[label=(\roman*)]
  \item one of 8 group activity labels, %
  \item one of 9 action labels per player, and%
  \item a bounding box per player. 
\end{enumerate*}
Bounding box tracklets of players along the 10 frames before and after the center frame are also provided by Bagautdinov~\etal~\cite{bagautdinov2017social}, and serve as the ground-truth bounding box labels of these frames. 
However, in the WSGAR setting, models including ours utilize the group activity labels only and disuse the stronger and fine-grained annotations.
We adopt Multi-class Classification Accuracy (MCA) and Merged MCA for evaluation throughout our experiments. 
In particular, for computing Merged MCA, we merge the classes \textit{right set} and \textit{right pass} into \textit{right pass-set}, and \textit{left set} and \textit{left pass} into \textit{left pass-set} as in SAM~\cite{yan2020social} for a fair comparison. 

\noindent
\textbf{NBA dataset.} 
This dataset consists of 7624 clips for training and 1548 clips for testing. 
Currently, it is the only dataset proposed for WSGAR, which only provides one of 9 group activity labels 
for each clip. 
Thanks to its low annotation cost, it is currently the largest group activity recognition dataset. 
Since each video clip is 6-second long and 
usually exhibits a nontrivial temporal structure,
the dataset requires a model that captures long-term temporal dynamics compared with other GAR benchmarks. 
Also, it is a challenging benchmark due to fast movement, camera view change, and a varying number of people in each frame. 
For evaluation, we adopt Multi-class Classification Accuracy (MCA) and Mean Per Class Accuracy (MPCA) metrics; MPCA is adopted due to the class imbalance issue of the dataset. 

\subsection{Implementation details}
\noindent
\textbf{Sampling strategy. }
For both datasets, $T$ frames are sampled using the segment-based sampling~\cite{wang2016temporal} 
and each frame is resized to $720 \times 1280$. Note that $T=18$ for the NBA dataset and $T=5$ for the Volleyball dataset. 

\noindent
\textbf{Hyperparameters.}
We adopt an ImageNet pretrained ResNet-18~\cite{resnet} as the backbone. 
For motion-augmented feature extraction, a $1 \times 1$ convolution operation reduces the channel dimension to $C^{\prime}=64$ and local neighborhood size is set to $P=11$. 
We stack 6 Transformer encoder layers with 4 attention heads and 256 channels for the NBA dataset, and 2 Transformer encoder layers with 2 attention heads and 256 channels for the Volleyball dataset. 
We test different numbers of learnable tokens including $K=1, 2, 4, 8, 12, 16$, and use 12 for both datasets.
For partial context aggregation module, three 1D convolutional layers with kernel size of 5 without padding are used for NBA and two 1D convolutional layers with kernel size of 3 with zero-padding are used for Volleyball. 
In the multi-head self-attention (MHSA) aggregation, a single layer MHSA with 256 channels is used for both datasets and the number of heads is 4 for NBA and 2 for Volleyball. 

\noindent
\textbf{Training.}
On both datasets, our model is optimized by ADAM~\cite{Adamsolver} with $\beta_{1}=0.9$, $\beta_{2}=0.999$,
and $\epsilon=\expnum{1}{8}$ %
for 30 epochs. 
Weight decay is set to 
$\expnum{1}{4}$ %
for the NBA dataset and 
$\expnum{1}{3}$ %
for the Volleyball dataset. 
Learning rate is initially set to $\expnum{1}{6}$ with linear warmup to $\expnum{1}{4}$ for 5 epochs, and linearly decayed after the $6^\textrm{th}$ epoch. 
We use a mini-batch of size 4 on NBA and 8 on Volleyball. 

\subsection{Comparison with the state-of-the-art methods}

\noindent
\textbf{NBA dataset.}
For the NBA dataset, 
we compare our method with the state of the art in GAR and WSGAR, which use bounding box proposals provided by SAM~\cite{yan2020social}, and also with recent video backbones in the weakly supervised learning setting. 
For a fair comparison, we set their backbones to ResNet-18 except for VideoSwin~\cite{liu2021video} and use only RGB frames as input for all the methods. 
Table~\ref{table:SOTA_NBA} summarizes the results.
Note that the scores of reproduced SAM~\cite{yan2020social} are higher than those reported in its original paper.
The proposed method beats all the GAR and WSGAR methods by a large margin: 14.2\%p of MCA and 14.4\%p of MPCA. 
Regarding complexity, our method demands less parameters and slightly more FLOPs than other GAR methods although we do not count the computational complexity of their object detectors. 
Our method is also compared with recent video backbones, ResNet-18 TSM~\cite{lin2019tsm} and VideoSwin-T~\cite{liu2021video}, used in conventional action recognition. 
Although these powerful backbones perform well in WSGAR, ours achieves the best. 
We also show the result of our method without the motion feature module, which still outperforms all the other methods.

\begin{table}[!t]
\begin{center}
\begin{tabular}{>{\arraybackslash}m{2.35cm}>{\centering\arraybackslash}m{1.3cm}>{\centering\arraybackslash}m{0.85cm}>{\centering\arraybackslash}m{0.8cm}>{\centering\arraybackslash}m{0.9cm}}
\hline
Method                                                      & \# Params     & FLOPs     & MCA       & MPCA      \\
\hline
\addlinespace[0.5ex]
\multicolumn{3}{l}{\textbf{Video backbone}} \\
\addlinespace[0.5ex]
TSM~\cite{lin2019tsm}                                       & 11.2M         & 303G      & 66.6      & 60.3      \\
VideoSwin~\cite{liu2021video}                               & 27.9M         & 478G      & 64.3      & 60.6      \\
\hline
\addlinespace[0.5ex]
\multicolumn{3}{l}{\textbf{GAR model}} \\
\addlinespace[0.5ex]
ARG~\cite{wu2019learning}                                   & 49.5M         & 307G      & 59.0      & 56.8      \\
AT~\cite{gavrilyuk2020actor}                                & 29.6M         & 305G      & 47.1      & 41.5      \\
SACRF~\cite{pramono2020empowering}                          & 53.7M         & 339G      & 56.3      & 52.8      \\
DIN~\cite{yuan2021spatio}                                   & 26.0M         & 304G      & 61.6      & 56.0      \\
{}\textsuperscript{\textdagger}SAM~\cite{yan2020social}     & -             & -         & 49.1      & 47.5      \\
SAM~\cite{yan2020social}                                    & 25.5M         & 304G      & 54.3      & 51.5      \\
\rowcolor{Gray}
Ours w/o motion                                             & 17.3M         & 311G      & \underline{73.6}  & \underline{69.0}  \\
\rowcolor{Gray}
Ours                                                        & 17.5M         & 313G      & \textbf{75.8}     & \textbf{71.2}     \\
\hline
\end{tabular}
\end{center}
\vspace{-1em}
\caption{Comparison with the state-of-the-art GAR models and video backbones on the NBA dataset. 
All models except VideoSwin adopt ResNet-18 backbone.
Numbers in \textbf{bold} indicate the best performance and \underline{underlined} ones are the second best. `{}\textsuperscript{\textdagger}' indicates that the result is copied directly from SAM~\cite{yan2020social}. 
All the other results are reproduced by us. 
}
\label{table:SOTA_NBA}
\vspace{-0.5em}
\end{table}

\begin{table}[!t]
\begin{center}
\begin{tabular}{>{\arraybackslash}m{2.4cm}>{\centering\arraybackslash}m{2.3cm}>{\centering\arraybackslash}m{0.9cm}>{\centering\arraybackslash}m{1.0cm}}

\hline
Method                             & Backbone              & MCA   & Merged MCA\\ [0.3ex]
\hline
\addlinespace[0.5ex]
\multicolumn{3}{l}{\textbf{Fully supervised}} \\
\addlinespace[1ex]
SSU~\cite{bagautdinov2017social}                 & Inception-v3          & 89.9  & - \\
PCTDM~\cite{yan2018participation}                & ResNet-18             & 90.3  & 94.3\\ %
StagNet~\cite{qi2018stagnet}                     & VGG-16                & 89.3  & - \\
ARG~\cite{wu2019learning}                        & ResNet-18             & 91.1 & \underline{95.1}\\ %
CRM~\cite{azar2019convolutional}        & I3D                   & 92.1  & - \\
HiGCIN~\cite{yan2020higcin}                      & ResNet-18             & 91.4  & - \\
AT~\cite{gavrilyuk2020actor}                     & ResNet-18             & 90.0 & 94.0\\ %
SACRF~\cite{pramono2020empowering}               & ResNet-18             & 90.7 & 92.7   \\ %
DIN~\cite{yuan2021spatio}                        & ResNet-18             & \underline{93.1}  & \textbf{95.6} \\ %
TCE+STBiP~\cite{yuan2021learning}        & VGG-16                & \textbf{94.1}  & - \\
GroupFormer~\cite{li2021groupformer}                      & Inception-v3             & \textbf{94.1}  & - \\

\addlinespace[1ex]\hline
\addlinespace[0.5ex]
\multicolumn{3}{l}{\textbf{Weakly supervised}} \\
\addlinespace[1ex]
PCTDM~\cite{yan2018participation}               & ResNet-18             & 80.5  & 90.0\\
ARG~\cite{wu2019learning}                       & ResNet-18             & 87.4  & 92.9\\
AT~\cite{gavrilyuk2020actor}                    & ResBet-18             & 84.3  & 89.6\\
SACRF~\cite{pramono2020empowering}              & ResNet-18             & 83.3  & 86.1  \\
DIN~\cite{yuan2021spatio}                       & ResNet-18             & 86.5  & 93.1\\
SAM~\cite{yan2020social}                        & ResNet-18             & 86.3  & 93.1\\
{}\textsuperscript{\textdagger}SAM~\cite{yan2020social}                        & Inception-v3          & -     & \underline{94.0}\\
\rowcolor{Gray}
Ours w/o motion                                 & ResNet-18             & \underline{88.1}  & \underline{94.0}\\
\rowcolor{Gray}
Ours                                            & ResNet-18             & \textbf{90.5}  & \textbf{94.4}\\
\hline
\end{tabular}
\end{center}

\vspace{-1.0em}
\caption{Comparison with the state-of-the-art methods on the Volleyball dataset. `-' indicates that the result is not provided, and `{}\textsuperscript{\textdagger}' indicates that the result is copied directly from SAM~\cite{yan2020social}. } 
\label{table:SOTA_Volleyball}
\vspace{-1.0em}
\end{table}

\noindent
\textbf{Volleyball dataset.}
For the Volleyball dataset, we compare our method with the state-of-the-art GAR and WSGAR methods in two settings: fully supervised setting and weakly supervised setting. 
The difference of two settings is the use of actor-level labels including ground-truth bounding boxes and individual action class labels, in both training and inference.
For a fair comparison, we report the results of previous methods~\cite{azar2019convolutional, yuan2021learning, li2021groupformer, yan2020higcin, bagautdinov2017social, qi2018stagnet} using only RGB inputs, and the reproduced results~\cite{yan2018participation, wu2019learning, gavrilyuk2020actor, pramono2020empowering, yuan2021spatio} using the ResNet-18 backbone. Note that the former is brought from the original papers and the latter is the MCA values from~\cite{yuan2021spatio}.
For the weakly supervised setting, we replace the ground-truth bounding boxes with an object detector pretrained on an external dataset and remove the individual action classification head. 
Table~\ref{table:SOTA_Volleyball} summarizes the results.
The first and second section show the results of previous methods in fully supervised setting and weakly supervised setting, respectively. 
Our method outperforms all the GAR and WSGAR models in weakly supervised setting by a substantial margin: 3.1\%p of MCA and 1.3\%p of Merged MCA when compared to the models %
using ResNet-18 backbone. 
It also beats the current state of the art based on Inception-v3.
We state the results without motion feature module to show its competitiveness.
Compared to the GAR methods in the fully supervised settings, our method surpasses the recent GAR methods~\cite{bagautdinov2017social, yan2018participation, qi2018stagnet, gavrilyuk2020actor, pramono2020empowering}, using stronger actor-level supervision.

\subsection{Ablation studies}

We also demonstrate the effectiveness of our method by ablation studies and analysis on the NBA dataset. 

\noindent
\textbf{Effects of the proposed modules.}
Table~\ref{tab:ablation_module} summarizes the effects of each module. 
Base model consists of a backbone, a global average pooling layer, and a group activity classifier.
Without the partial context aggregation (PCA), all token embeddings are averaged to form a group representation. 
Unless the motion feature module is inserted, the original backbone is utilized.
All training setups except the model architecture are same. 
From the result, all three components consistently enhance the model in terms of two metrics.
Partial context embedding (PCE) module improves MCA from 58.4\% to 64.1\% and MPCA from 51.7\% to 58.5\%. 
Regarding that the base model predicts a group activity using a global scene feature, it emphasizes the importance of capturing partial contexts for recognizing a group activity. 
PCA module further increases MCA from 64.1\% to 73.6\% and MPCA from 58.5\% to 69.0\%.
This also shows the effectiveness of our aggregation method versus naive average pooling aggregation. 
The detailed analysis of the token aggregation methods will be discussed later (Table~\ref{tab:ablation_aggregation}). 
Motion feature module gives an additional gain regardless of other components, which implies that the sense of motion helps to understand group activities.

\noindent
\textbf{Effects of the token aggregation methods.}
For this ablation, we do not adopt motion feature module, and plain ResNet-18 backbone is used as a feature extractor. 
Table~\ref{tab:ablation_aggregation} shows the performance of various token aggregation methods. 
The first and second section of the table contains the results of one-stage aggregation and two-stage aggregation method, respectively. 
Given a bag of token embeddings, the two-stage aggregation method separates the aggregation across different frames and the aggregation across different tokens, whereas the one-stage aggregation methods fuse all tokens at once.
MLP aggregation concatenates token embeddings and project them using a linear layer, and it is regarded as a baseline for aggregating vectors along certain axis.
Multi-head self-attention (MHSA) aggregation is composed of a single layer MHSA and an average pooling layer. 
From the result, the two-stage aggregation methods exceed the one-stage methods in most cases. 
Compared to the two-stage MLP aggregation, our method, 1D convolution across frames followed by MHSA across tokens, performs better. 
This shows the effectiveness of temporal convolution and MHSA aggregation, which are robust to a temporal shift in the video and able to capture dynamic relation between partial contexts. 
In addition, the order of aggregation demonstrates notable difference: Assembling with respect to temporal axis first and then to token axis brings a gain of 1.2\%p of MCA and 2.0\%p of MPCA.

\begin{table}[!t]
\begin{center}
\begin{tabular}{lcc}
\hline
Model                       & MCA               & MPCA \\
\hline
Base model                  & 58.4              & 51.7 \\
Base model + Motion         & 62.7              & 55.1 \\
\hline

PCE                         & 64.1              & 58.5 \\
PCE + Motion                & 65.1              & 59.1 \\
PCE + PCA                   & 73.6              & 69.0 \\
\rowcolor{Gray}
PCE + PCA + Motion          & $\textbf{75.8}$   & $\textbf{71.2}$ \\
\hline
\end{tabular}
\end{center}
\vspace{-1.0em}
\caption{Contributions of the proposed modules. 
PCE and PCA denote partial context embedding and partial context aggregation modules, respectively. 
}
\label{tab:ablation_module}
\vspace{-0.5em}
\end{table}

\begin{table}[!t]
\begin{center}
\begin{tabular}{lcc}
\hline
Aggregation method                  & MCA               & MPCA      \\
\hline
Average pooling                     & 64.1              & 58.5    \\
Max pooling                         & 61.3              & 55.1    \\
MLP (both $T$, $K$- dim)            & 62.8              & 56.4    \\
MHSA (both $T$, $K$- dim)           & 68.7              & 64.7    \\
\hline
MLP ($T$- dim) - MLP ($K$- dim)     & 68.9              & 63.1      \\
MHSA ($K$- dim) - 1D conv ($T$- dim) & 72.4             & 67.0      \\
\rowcolor{Gray}
Ours (1D conv - MHSA)               & \textbf{73.6}     & \textbf{69.0}     \\
\hline
\end{tabular}
\end{center}
\vspace{-1em}
\caption{Ablation on the token aggregation methods. $T$-dim and $K$-dim imply aggregation with the same token at different frames and aggregation with different tokens at the same frame, respectively. MHSA stands for multi-head self-attention. }
\label{tab:ablation_aggregation}
\vspace{-0.5em}
\end{table}

\begin{figure}
\begin{minipage}{\linewidth}
\begin{minipage}[c]{0.4\linewidth}
\vspace{1.3em}
\begin{tabular}[c]{>{\centering\arraybackslash}m{1.1cm}>{\centering\arraybackslash}m{0.8cm}>{\centering\arraybackslash}m{0.8cm}}
    \hline
    \# token    & MCA               & MPCA              \\
    \hline
    1           & 72.7              & 66.6              \\
    2           & 73.1              & 67.2              \\
    4           & 72.6              & 67.7              \\
    8           & 74.0              & \textbf{69.7}  \\
    12          & 73.6              & 69.0              \\
    16          & \textbf{74.3}     & 68.8              \\
    \hline
\end{tabular}
\captionof{table}{Ablation on the number of tokens per frame. }
\label{tab:ablation_num_token}
\end{minipage}
\hfill
\begin{minipage}[c]{0.5\linewidth}
\begin{tabular}[c]{>{\centering\arraybackslash}m{1.1cm}>{\centering\arraybackslash}m{0.8cm}>{\centering\arraybackslash}m{0.8cm}}
    \hline
    Model                   & MCA                   & MPCA  \\
    \hline
    Base                    & 73.6                  & 69.0  \\
    \textit{$res_{2}$}      & 72.2                  & 68.3  \\
    \textit{$res_{3}$}      & 74.2                  & 70.2  \\
    \textit{$res_{4}$}      & 74.8                  & 69.4  \\
    \textit{$res_{5}$}      & 74.0                  & 70.5  \\
    \textit{$res_{3,4}$}    & 74.2                  & 69.0  \\
    \textit{$res_{4,5}$}    & \textbf{75.8}         & \textbf{71.2}  \\
    \textit{$res_{3,4,5}$}  & 73.1                  & 68.1  \\
    \hline
\end{tabular}
\captionof{table}{Ablation on the position of the motion feature module. Base represents the model without the module. }
\label{tab:ablation_motion}
\end{minipage}
\end{minipage}
\vspace{-1.5em}
\end{figure}

\begin{figure*}[!t]
\begin{center}
\includegraphics[width=1.0\linewidth]{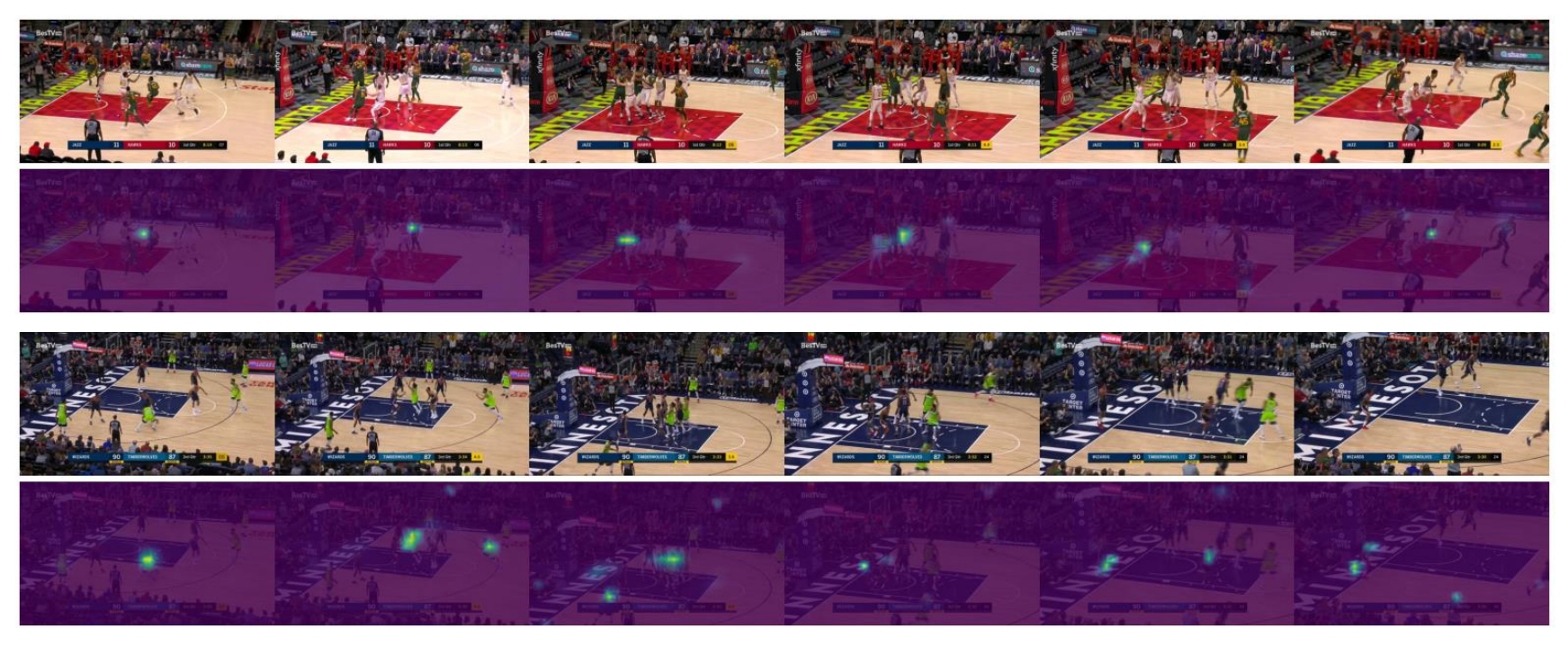}
\end{center}
\vspace{-2.0em}
\caption{Visualization of the Transformer encoder attention maps on the NBA dataset. }
\label{fig:fig4}
\vspace{-1.0em}
\end{figure*}

\begin{figure*}[!t]
\begin{center}
\includegraphics[width=1.0\linewidth]{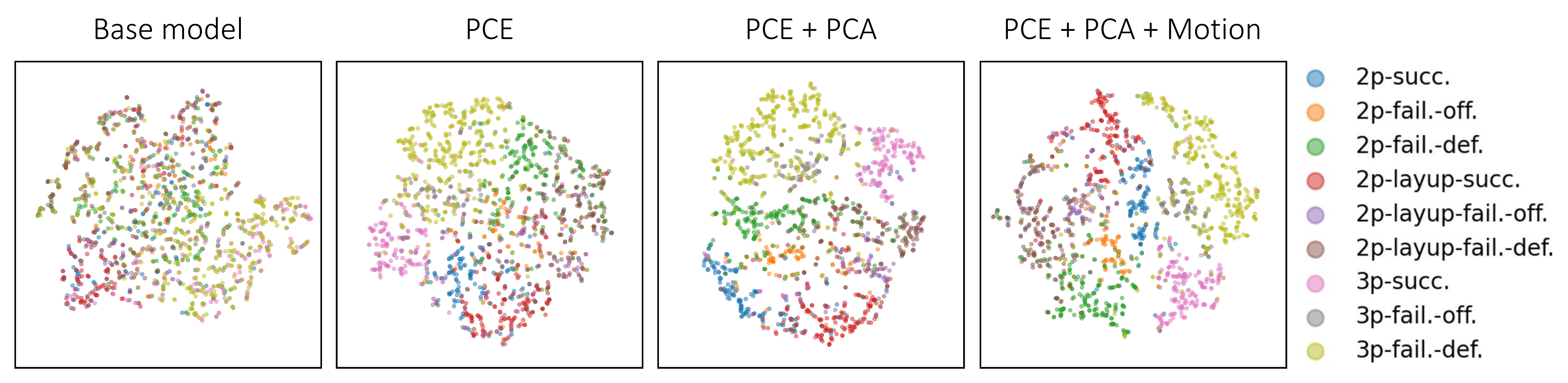}
\end{center}
\vspace{-1.5em}
\caption{$t$-SNE~\cite{tsne} visualization of feature embedding learned by different model variants on the NBA dataset. }
\label{fig:fig5}
\end{figure*}

\noindent
\textbf{Effects of the number of tokens. }
Table~\ref{tab:ablation_num_token} summarizes the performance of different number of tokens. 
Note that the ablation is conducted without the motion feature module.
When the number of learnable tokens is set to 1, a single token embedding extracts context information as a global scene feature. 
The performance tends to increase with the number of tokens in general, and the optimal result is gained when the number is in the range of 8 to 16.
These results indicate that it is more effective to divide the scene context into several pieces in order to encapsulate enriched group representation.

\noindent
\textbf{Effects of the position of the motion feature module. }
We investigate the efficacy of inserting the motion feature module at different positions. 
The motion feature is placed after a residual block and it is denoted as $res_{i}$ if the motion feature module is inserted after the $i$-th residual block. 
As shown in Table~\ref{tab:ablation_motion}, the motion feature module is effective in most cases except the case where it is inserted after the 2nd residual block, which is considered too early to compute the local correlation. 
Putting multiple motion feature modules is also tested, and inserting motion feature modules after 4th and 5th block performs the best. 

\subsection{Qualitative analysis}
In Fig.~\ref{fig:fig4}, we illustrate the attention visualizations obtained from the final Transformer encoder layer on the NBA dataset. 
The results imply that token embeddings are learned to attend key concepts and follow the activity occurs in a given video clip. 
Fig.~\ref{fig:fig5} displays the $t$-SNE~\cite{tsne} visualization results of our model and its variants. 
Final group representation of each model on NBA is visualized in two-dimensional space. 
We can find that each proposed modules contribute to a clear separation of each class.

%% file: _5_conclusion_arxiv.tex
\section{Conclusion}
We have presented a detector-free method for weakly supervised group activity recognition, which first embeds the partial contexts of an activity through the attention mechanism, then aggregates them while capturing their temporal evolution. 
We achieve the state of the art on two benchmarks in weakly supervised learning setting, and even outperform several GAR models that rely on stronger supervision. 
These results suggest that partial contexts captured by our method could be more effective than the human prior given in the form of person bounding boxes.
Albeit with such benefits, our model has difficulties generating sufficiently diverse token embeddings due to the absence of direct supervision. 
Further improvement could be achieved by increasing their diversity without capturing irrelevant contexts.

%% file: _supp_1_experiments_arxiv.tex
\section{Experimental details}

\subsection{Implementation of reproduction}

\noindent
\textbf{NBA dataset. }
We reproduce GAR~\cite{wu2019learning, gavrilyuk2020actor, pramono2020empowering, yuan2021spatio} and WSGAR~\cite{yan2020social} methods following the official code of DIN\footnote{\label{note1}Original DIN codes are available at \url{https://github.com/JacobYuan7/DIN_GAR}.}~\cite{yuan2021spatio} and the implementation description illustrated in its original paper, repectively. 
For a fair comparison, segment-based sampling~\cite{wang2016temporal}, batch size of 4, the number of bounding boxes $N=12$, and the number of frames $T=18$ are applied for all methods.
The only difference from the implementation of DIN is that all methods are trained in an end-to-end manner. 
Unless specified, all other hyperparameters are identical to those of the code from DIN.
We provide more implementation details of each model below. 

\begin{itemize}
\itemsep=-0.5mm
    \item \textbf{ARG}~\cite{wu2019learning}
    ResNet-18 backbone replaces the original backbone of Inception-v3. 
    \item \textbf{AT}~\cite{gavrilyuk2020actor}
    A single RGB branch is utilized and ResNet-18 backbone replaces the backbone of I3D and HRNet. 
    \item \textbf{SACRF}~\cite{pramono2020empowering}
    The backbone is replaced to ResNet-18, and its multiple modalities are substituted to single RGB input. 
    Since NBA dataset does not have individual action labels, we remove the unary energy term. 
    \item \textbf{DIN}~\cite{yuan2021spatio}
    The experiment is conducted following its official implementation. 
    \item \textbf{SAM}~\cite{yan2020social}
    The number of proposals ($N^p$) and the number of selected proposals ($K^p$) are set to 14 and 8, respectively.
\end{itemize}

\noindent
\textbf{Volleyball dataset. }
Reproduction of the following models is also based on the code of DIN\textsuperscript{\ref{note1}}~\cite{yuan2021spatio}.
Each reproduction first goes through backbone training process regarding the number of categories, then proceeds to train each inference module afterward.
Note that MCA values of the following models under fully supervised setting are brought from DIN~\cite{yuan2021spatio}, hence we provide implementation detail of experiments under \textbf{A)} fully supervised setting with the aim of classifying actions into 6 (merged) labels, \textbf{B)} weakly supervised setting/8 labels, and \textbf{C)} weakly supervised setting/6 labels.
In weakly supervised setting, actor bounding boxes are replaced to proposal boxes generated by Faster R-CNN~\cite{faster_rcnn} pretrained on COCO dataset~\cite{coco} and individual action annotations are eliminated.
In general, we use a batch size of 2 and the number of frames $T = 10$ for the following work.
Unless mentioned, other hyperparameters are set based on the code provided by DIN.
Followings are further implementation details of each model.

\begin{itemize}
\itemsep=-0.5mm
\item \textbf{PCTDM}~\cite{yan2018participation}
ResNet-18 is applied instead of AlexNet, and RoIAlign features replace cropped/resized individual images of the original paper. 
Furthermore, weight decay rate of $1 \times 10^{-4}$ is applied to \textbf{C)}.
\item \textbf{ARG}~\cite{wu2019learning} 
Likewise, its backbone is changed to ResNet-18. Unlike its original setting, the backbone training is allowed in the model training process for a fair comparison with other models.

\item \textbf{AT}~\cite{gavrilyuk2020actor}
A single RGB branch is utilized and ResNet-18 backbone replaces the backbone of I3D and HRNet.

\item \textbf{SACRF}~\cite{pramono2020empowering}
The backbone is replaced to ResNet-18, and its multiple modalities are substituted to single RGB input.
Due to the removal of individual action labels, the unary energy term is removed.

\item \textbf{DIN}~\cite{yuan2021spatio}
The experiment is conducted following its official implementation.

\end{itemize}
\begin{figure*}
     \centering
     \begin{subfigure}[b]{0.31\textwidth}
         \centering
         \includegraphics[width=\textwidth]{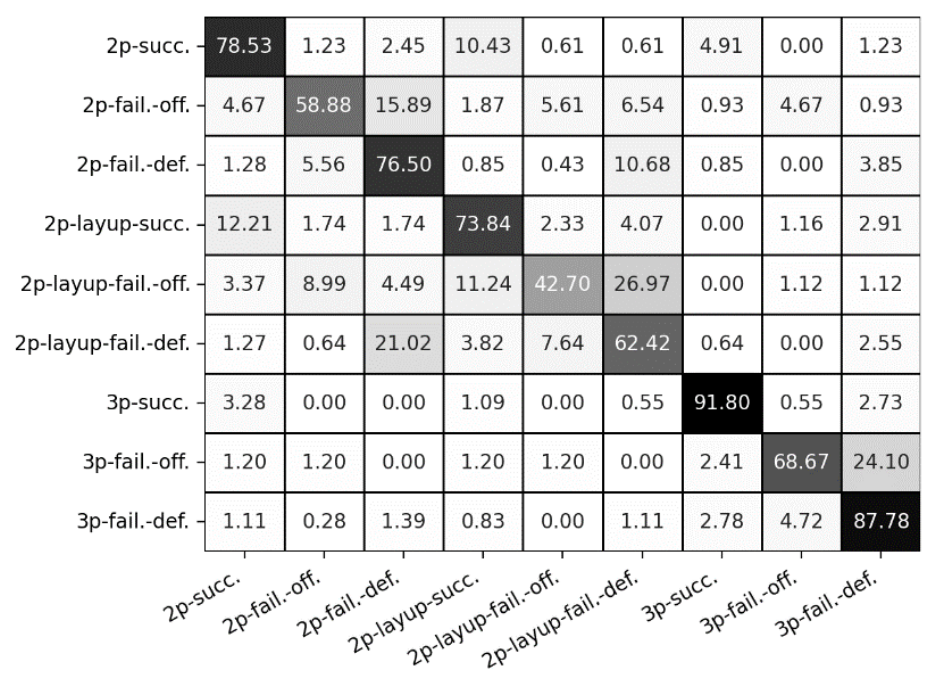}
         \caption{}
         \label{fig:confusion_NBA}
     \end{subfigure} \hspace{2mm}
     \begin{subfigure}[b]{0.31\textwidth}
         \centering
         \includegraphics[width=\textwidth]{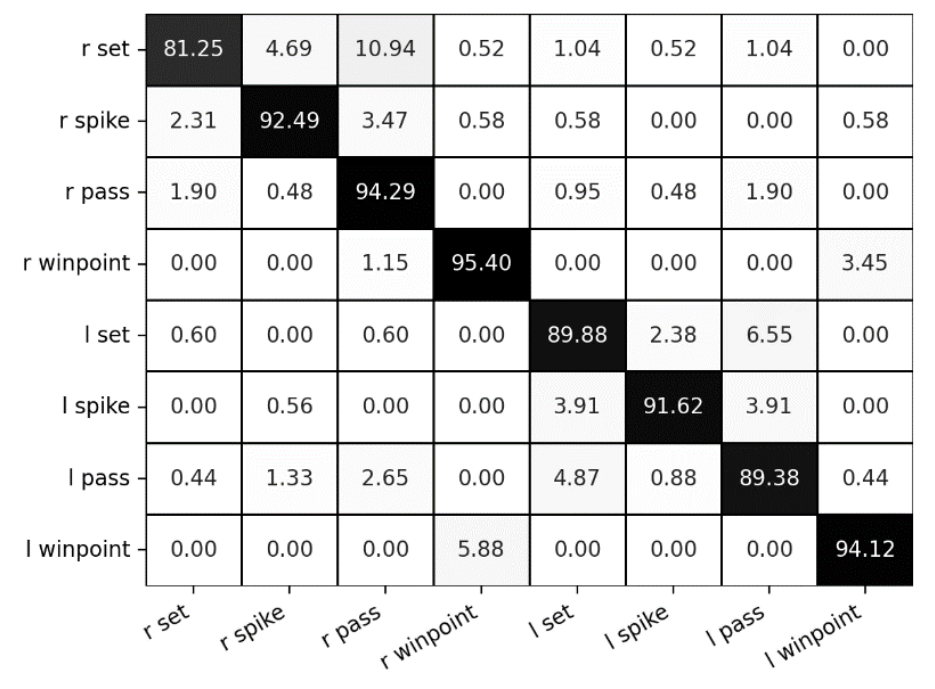}
         \caption{}
         \label{fig:confusion_MCA}
     \end{subfigure} \hspace{2mm}
     \begin{subfigure}[b]{0.31\textwidth}
         \centering
         \includegraphics[width=\textwidth]{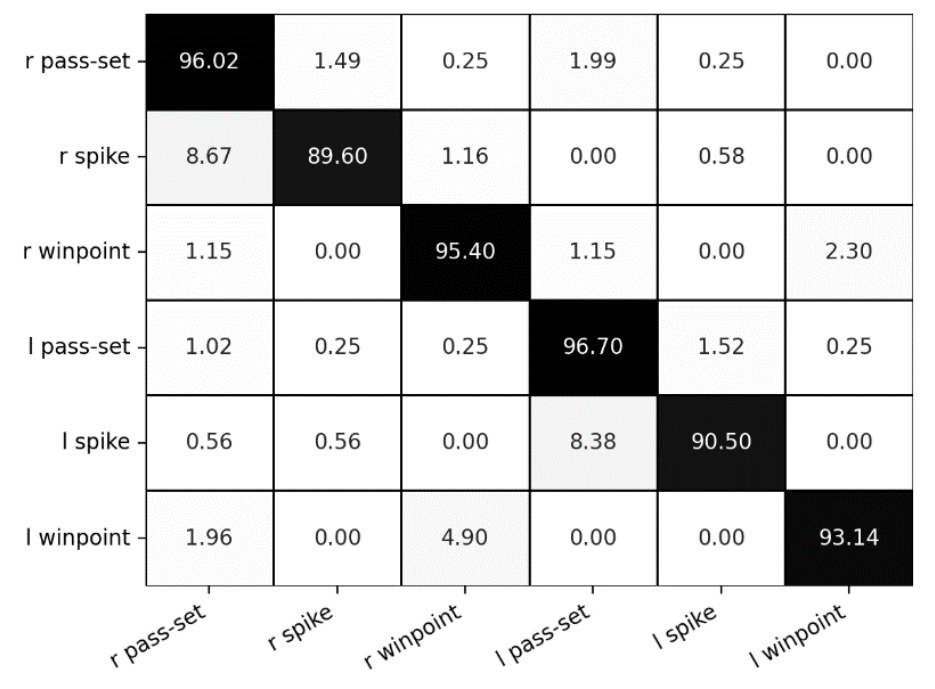}
         \caption{}
         \label{fig:confusion_MPCA}
     \end{subfigure}
    \caption{
    The confusion matrix 
    (a) on the NBA dataset, 
    (b) of the original 8 class classification on the Volleyball dataset, and
    (c) of the merged 6 class classification (merge \textit{pass-set} class) on the Volleyball dataset. 
    }
     \label{fig:confusion}
\end{figure*}

SAM~\cite{yan2020social} is reproduced following the method described in the original paper.
The major difference with DIN-based reproductions is that it occupies a batch size of 8, a dropout rate of 0.1, and $T=3$ frames. 

\begin{itemize}
\item \textbf{SAM}~\cite{yan2020social}
Note that SAM itself is a WSGAR work, so \textbf{A)} is disregarded. Since \textbf{C)} is already conducted in the original paper, we only reproduce \textbf{B)}. The number of proposals ($N^p$) and the number of selected proposals ($K^p$) are set to 16 and 12, respectively. 
\end{itemize}

\subsection{Implementation of video backbones}
We reproduce recent video backbones, ResNet-18 TSM~\cite{lin2019tsm} and VideoSwin-T~\cite{liu2021video} following the official codes. 
For a fair comparison, sampling strategy and training details are the same as ours.

\subsection{Motion-augmented backbone}

We use ResNet-18~\cite{resnet} backbone in our experiment. 
We provide details of the backbone architectures to understand to which place the motion feature modules are inserted. 
Table~\ref{table:ResNet_detail} shows the ResNet-18 backbone architectures. 
For NBA dataset, we insert two motion feature modules after 4th and 5th residual block. 
For Volleyball dataset, we insert one motion feature module after the last residual block.

\section{More ablation studies}

In this section, we provide additional ablation on NBA dataset. 
Note that we do not adopt motion feature computation module in this additional ablations and use plain ResNet-18 backbone as a feature extractor. 

\noindent
\textbf{Effects of the temporal convolution layers. }
Table~\ref{table:ablation_temp_conv} summarizes the performance according to different numbers and kernel sizes of temporal convolution layers. 
Note that we do not utilize zero-padding in this experiment. 
In most cases, MCA and MPCA increase as 1D convolutional layers are stacked. 
The performance is also affected by the kernel size of 1D convolutional layers, and it increases as the receptive field of temporal convolution layers gets wider.

\begin{table}[!t]
\begin{center}
\begin{tabular}{c|c|c}
\hline
Layers                  & ResNet-18                                                                     & Feature map size \\
\hline
\textit{$conv_{1}$}     & $7 \times 7$, 64, stride (2, 2)                                               & $T \times 360 \times 640$ \\
\hline
\textit{$pool_{1}$}     & $3 \times 3$, stride (2, 2)                                                   & $T \times 180 \times 320$ \\
\hline
\textit{$res_{2}$}      & \bsplitcell{ $3 \times 3$, 64 \\ $3 \times 3$, 64} $\times 2$                 & $T \times 180 \times 320$ \\
\hline
\textit{$res_{3}$}      & \bsplitcell{$3 \times 3$, 128 \\ $3 \times 3$, 128} $\times 2$                & $T \times 90 \times 160$ \\
\hline
\textit{$res_{4}$}      & \bsplitcell{$3 \times 3$, 256 \\ $3 \times 3$, 256} $\times 2$                & $T \times 45 \times 80$ \\
\hline
\textit{$res_{5}$}      & \bsplitcell{$3 \times 3$, 512 \\ $3 \times 3$, 512} $\times 2$                & $T \times 23 \times 40$ \\
\hline
\end{tabular}
\end{center}
\caption{ResNet-18 backbone details. $[k \times k, c] \times n$ denotes $n$ convolutional layers with kernel size of $k$ and $c$ channels. }
\label{table:ResNet_detail}
\end{table}

\begin{table}[!t]
\begin{center}
\begin{tabular}{cccc}
\hline
Model                                   & \# params         & MCA               & MPCA  \\
\hline
$[3 \times 1$, 256$] \times 3$          & 16.91M            & 72.7              & 68.0  \\
$[5 \times 1$, 256$] \times 1$          & 16.65M            & 70.2              & 65.5  \\
$[5 \times 1$, 256$] \times 2$          & 16.98M            & 71.1              & 65.2  \\
\hline
\rowcolor{Gray}
Ours ($[5 \times 1$, 256$] \times 3$)   & 17.31M            & \textbf{73.6}     & \textbf{69.0}  \\
\hline
\end{tabular}
\end{center}
\caption{Ablation on the different forms of temporal convolution layers. $[t \times 1, D] \times n$ denotes $n$ 1D convolutional layers with kernel size of $t$ and $D$ channels. }
\label{table:ablation_temp_conv}
\end{table}

%% file: _supp_2_qualitative_arxiv.tex
\section{More experimental results}

In this section, we provide more visualizations and qualitative results that are omitted in the main paper due to the space limit. 

Fig.~\ref{fig:confusion} shows the confusion matrix on NBA and Volleyball datasets. 
For the NBA dataset (Fig.~\ref{fig:confusion_NBA}), the most confusing cases are \textit{2p-layup-fail.-off.} versus \textit{2p-layup-fail.-def.} which only differs who rebound a ball after shooting a layup. 
For the 8 class classification (Fig.~\ref{fig:confusion_MCA}), the most confusing cases are \textit{r set} versus \textit{r pass} and \textit{l set} versus \textit{l pass}. 
This is challenging because our model does not utilize individual action label in training which gives clues to classify \textit{set} and \textit{pass} class. 
For the merged 6 class classification (Fig.~\ref{fig:confusion_MPCA}) which merges \textit{pass} and \textit{set} class into \textit{pass-set} class, our model achieves satiable accuracies on \textit{right pass-set} and \textit{left pass-set} class. 
Nevertheless, our model struggles to classify \textit{spike} and \textit{pass-set} due to a class imbalance problem. 

\begin{figure*}[!t] \vspace{20pt}
\begin{center}
\includegraphics[width=1.0\linewidth]{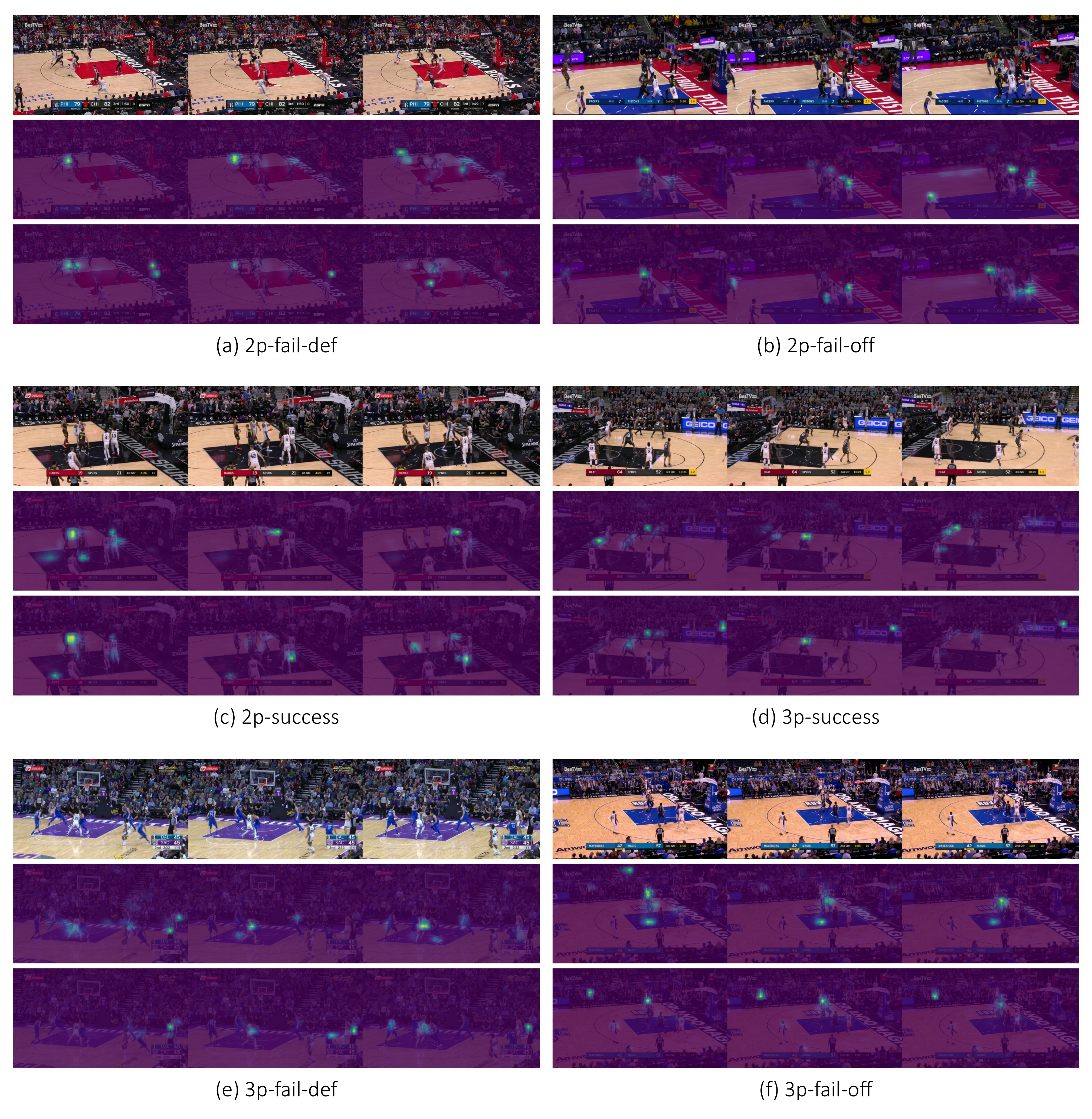}
\end{center}
\caption{Visualizations of the cross-attention maps on NBA dataset. Attention maps for 2 tokens among 12 tokens are displayed. Tokens tend to capture different part of each group activity: in this example, the first token focuses more on activities happening among players and the ball, while the second token weighs more on peripheral clues besides the main scene. \vspace{60pt}}

\label{fig:visualization_nba}
\end{figure*}

\begin{figure*}[!t] \vspace{20pt}
\begin{center}
\includegraphics[width=1.0\linewidth]{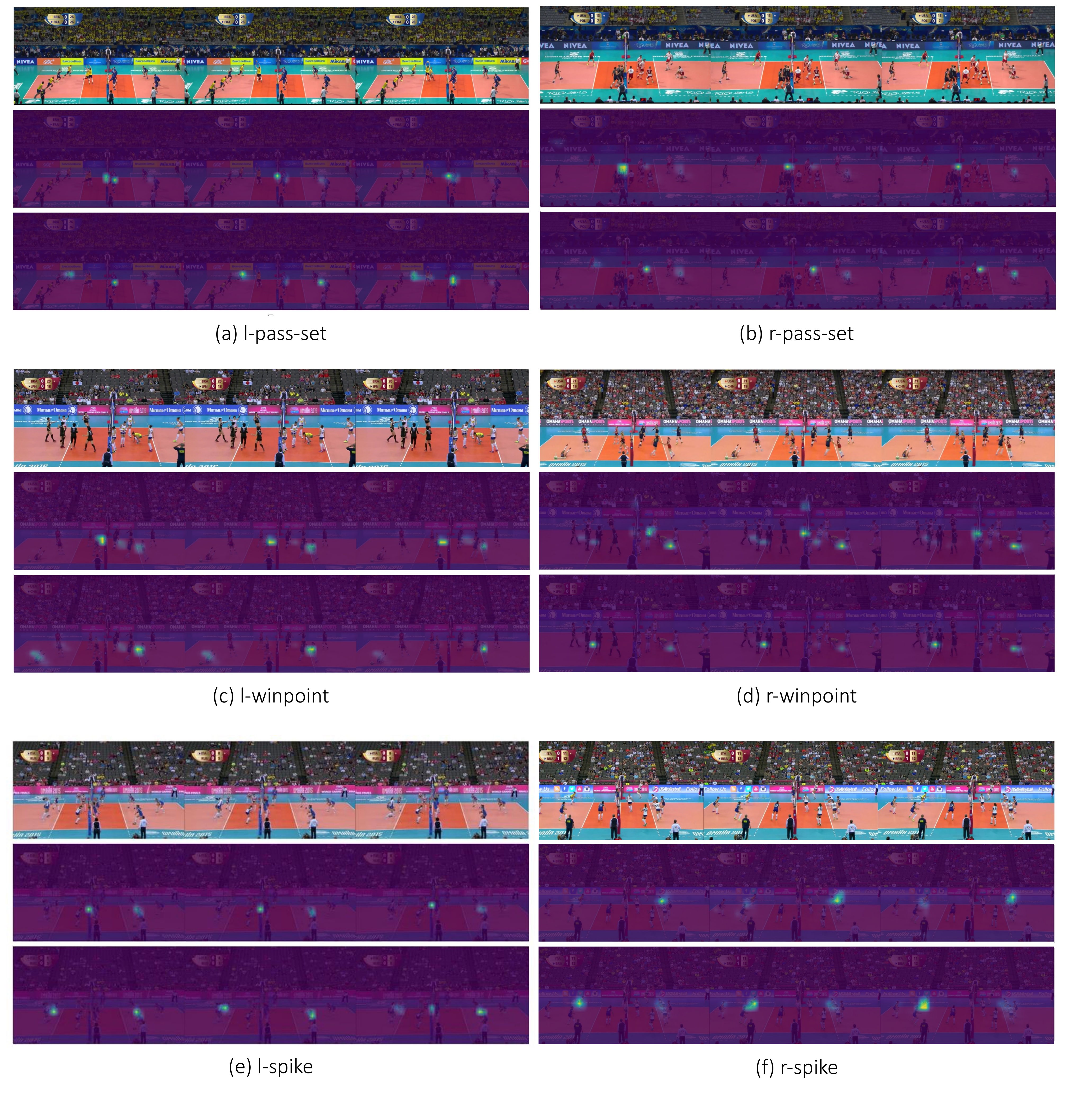}
\end{center}
\caption{Visualizations of the cross-attention maps on Volleyball dataset. Attention maps for 2 tokens among 12 tokens are displayed. Likewise, tokens understand given group activities in a partial way. The first token watches more on activities happening around the net, while the second token shows more tendency toward capturing activities that involve players and happen farther from the net. }

\label{fig:visualization_volleyball}

\end{figure*}

Fig.~\ref{fig:visualization_nba} and ~\ref{fig:visualization_volleyball} show more visualizations on NBA and Volleyball dataset, respectively.

